# Rotated Digit Recognition by Variational Autoencoders with Fixed Output Distributions

David Yevick
Department of Physics
University of Waterloo
Waterloo, ON N2L 3G7
yevick@uwaterloo.ca

**Abstract:** This paper demonstrates that a simple modification of the variational autoencoder (VAE) formalism enables the method to identify and classify rotated and distorted digits. In particular, the conventional objective (cost) function employed during the training process of a VAE both quantifies the agreement between the input and output data records and ensures that the latent space representation of the input data record is statistically generated with an appropriate mean and standard deviation. After training, simulated data realizations are generated by decoding appropriate latent space points. Since, however, standard VAE:s trained on randomly rotated MNIST digits cannot reliably distinguish between different digit classes since the rotated input data is effectively compared to a similarly rotated output data record. In contrast, an alternative implementation in which the objective function compares the output associated with each rotated digit to a corresponding fixed unreferenced reference digit is shown here to discriminate accurately among the rotated digits in latent space even when the dimension of the latent space is 2 or 3.

**1. Introduction:** Pattern recognition attempts to simulate the human ability to discern classes of objects that are highly distorted through geometric transformations or random shape variations. Perhaps the most established strategy employs sets of image feature vectors that possess small intra-class and large inter-class distances insuring, respectively, the robustness and discriminability of the classification process. [1,2] However, this approach is generally complicated, especially if the classification is required to be invariant under geometrical transformations. Non-orthogonal [3–5] and subsequently orthogonal [6,7] moments with corresponding invariants and basis functions were subsequently introduce which improved the discriminative capability and robustness with respect to random noise and facilitated the reconstruction of pattens from the basis function expansions. These basis functions are generally mutually orthogonal on the unit circle enabling the simple and reliable identification of rotation-invariant features. [8] In particular, harmonic function based circularly orthogonal moments [9–11] can generally more accurately describe sets of rotated images and further incur less computational overhead than procedures based on polynomial moments. [12] Furthermore, while most moment formulations are based on continuous analytic basis functions, discrete basis functions that are compatible with the image data have been advanced in order to alleviate numerical issues. [13–16] However, since the discrete orthogonal basis functions are analytically involved, the implementation of discrete moments and rotation invariants requires considerable computer resources in comparison to the continuous formulation. [17,18]

Automated procedures have also been developed that extract feature descriptors or classifiers from sufficiently representative datasets. The most common techniques employ deep learning which is effective for complete training sets. In contrast to standard numerical or analytic methods machine learning extracts features of a data set through multivariable optimization. The limited number of data records and the presence of computational metaparameters such as the number and connectivity of the network computing elements in the optimization procedure therefore introduce intrinsic errors. [19]

However, if the underlying system data is affected by stochasticity or measurement uncertainty, machine learning can yield predictions that are at least as accurate as those of deterministic procedures.

This paper presents an automated approach to character recognition based on a modification of the standard variational encoder that enables the identification of rotated characters without the analytic or programming overhead associated with the moment formalism or advanced deep learning methods. The modified formalism extends the domain of applicability of the variational autoencoder technique as is clearly evident from the MNIST results of the following section. Character recognition is a central application of pattern recognition that not only possesses numerous practical applications but also benchmarks the ability of a recognition system to classify objects independently of their positions, orientations or spatial extents. Character recognition methods that are invariant with respect to such transformations have been developed that employ statistical [20,21] as well as structural approaches based on Fourier descriptors [22], decomposition techniques [23] and the moment procedures cited above. Further, advanced neural network procedures applied to character recognition often exhibit improved performance but are architecturally complicated and accordingly resource intensive. For example, while high order neural networks can realize invariance, [24,25] the large number of neural connections that require training often limit the applicability of these techniques. Neural network procedures that intrinsically incorporate the desired invariances have accordingly been developed, as for example the double backpropagation algorithm. This technique employs a preprocessor to extract the geometrical features of the input pattern such that the features rotated or scaled patterns are cyclically shifted version of those of the standard pattern, after which an appropriate transform is employed to generate an invariant output. [26] Examples of other approaches are the Neocognitron method [27], which however is not particularly effective for rotated images, and a neural network that employs many layers of neurons, each of which is invariant with respect to a particular degree of rotation, scaling or translation, again necessitating a large number of connections. Rotational invariance can also be achieved in conjunction with convolutional neural networks, (CNN) for example by examining the histogram of the output of the convolutions which will be rotation and translation invariant[28] by randomly performing translations, scaling and rotation of the CNN feature map during learning[29] or by learning a rotation-invariant layer through a modification of objective function.[30]  A standard machine learning approach can also be applied to a deep network by employing as input a set of rotated and/or translated images, although this procedure is intrinsically costly.[31]

In contrast to the above approaches, this paper formulates the character or more generally pattern recognition problem in the simplified context of Variational Autoencoders (VAE:s). [32]  While supplanted to a certain extent been supplanted by more recent techniques such as generative adversarial networks, [33] VAE:s are frequently employed in deep learning contexts such as computer vision [34] to provide a generative model that efficiently produces simulated data with the same properties as the sample set. However, the simplicity and flexibility of the VAE has motivated continued interest in the method. In particular, the VAE, resembles principal component analysis in that it effectively compresses the information contained in the data into a set of parameters termed latent variables that is considerably smaller than the number of degrees of freedom of each data record. [35,36] The VAE architecture can be enhanced by incorporating, for example, different deep neural network architectures and optimized objective functions. [37,38]  Further, through the unsupervised learning of representations in which each latent variable maps to a particular physical order parameter or physical feature such as for example, gender, presence or absence of glasses or hair, etc., VAE:s enable interpretable latent code manipulation, which can be leveraged in many practical contexts. [36,39]

While VAE reconstruction is easily implemented and computationally efficient, the L2 (logistic regression) distance between the generated and original images is often significant, especially for latent spaces with few dimensions. This effect is clearly identified in simple models such as the benchmark Ising model. [39] However, it is manifested for images as a high degree of blur that is not necessarily evident from the pixel-by-pixel loss which can also be anomalously large if the reconstructed image is slightly rotated or offset. [40] Since the final layers of a deep CNN capture long-range spatial correlations in the input image rather than local pixel-to-pixel variations, feature perceptual loss functions that better quantify visual perception can be constructed by comparing the latent representations [41] or the VAE latent vectors [37] of two images within a VAE constructed from convolutional neural networks.

To illustrate in a straightforward and unambiguous fashion that invariances can be learned and subsequently employed to recognize characters and by extension patterns though simple modifications of the VAE procedure, this paper examines the latent space distributions generated by the MNIST digits after rotation through random angles between 0 and $2\pi$. Unlike non-rotated digits for which each value is mapped to a unique, well-defined region in the latent variable space, the latent spaces occupied by randomly rotated digits with different values overlap even for relatively high dimensional latent vector spaces. The results below, however, establish that the latent space can in fact be structured by employing as a metric for each rotated input digit belonging to a given class (e.g. numerical value) the deviation of the VAE output from a fixed unrotated digit that is representative of the digits in the class. The partitioning of the latent space among digits is then distinct even for two latent variables. To demonstrate further the validity and potential applicability of the method, the region in latent space in the vicinity of the latent space position corresponding to the reference pattern of a given digit is shown to map almost exclusively to different rotated versions of the digit. This implies that data can be reliably categorized by the procedure subject to the fidelity of the training set and the degree to which the reference patterns embody the significant distinguishing features of the object classes.

While this procedure resembles to a certain extent a simplified version of the feature loss function technique, it can be potentially generalized to more complex pattern recognition problems involving randomly rotated or transformed data. For example, a VAE could be trained on many input images by associating each of these with one of a set of unrotated standardized images with specific features such as hair or eye color. Subsequently a trained VAE would potentially associate a unique latent vector space volume with each of these standardized image features, enabling the essential content of a new image to be inferred from its latent space representation.

**2. Variational Autoencoders:**

*Standard Procedure*: As noted in the previous section, a variational encoder maps the underlying features of high-dimensional training data onto a lower dimensional space termed the latent vector space. Simulated data that reproduces and interpolates between the essential features of the training data set can then be generated from the latent space representation. To summarize the theoretical basis of the technique, a VAE learns the essential common features of related images of an input dataset $X = \{x^{(i)}, i = 1, 2, \ldots N\}$ consisting typically of $N$ independent (iid) image records. If the probability of a certain data realization is denoted $\tilde{p}(\vec{x})$ and the system is modelled by a set of parameters $\vec{\theta}$ with probability distribution $p_{\vec{\theta}}(\vec{x})$, the logarithm of the probability of the data under the model assumptions is

$$\log p_{\vec{\theta}}(X) = \sum_{\vec{x} \in X} \log p_{\vec{\theta}}(\vec{x}) \tag{1}$$

The latent hidden variables of the VAE, $\vec{z}$, are associated with a neural network which generates a joint distribution $p_{\vec{\theta}}(\vec{x}, \vec{z})$, that generally factorizes such that $p_{\vec{\theta}}(\vec{x}, \vec{z}) = p_{\vec{\theta}}(\vec{x}|\vec{z})P(\vec{z})$, where $P(\vec{z})$ is termed the prior distribution over $\vec{z}$. However, randomly chosen samples in $\vec{z}$ will map to points with small $p_{\vec{\theta}}(\vec{x})$ (the model evidence) which coincides with the marginalized distribution $p_{\vec{\theta}}(\vec{x}) = \int p_{\vec{\theta}}(\vec{x}, \vec{z}) d\vec{z}$. To restrict the latent space to statistically significant samples, the VAE employs a neural network encoder $q_{\vec{\phi}}(\vec{z}|\vec{x})$ that inverts the action of the decoder; that is $q_{\vec{\phi}}(\vec{z}|\vec{x}) \approx p_{\vec{\theta}}(\vec{z}|\vec{x})$. This encoder accordingly approximates the intractable posterior $p_{\theta}(\vec{z}|\vec{x})$ of the generative model. Applying the trained stochastic decoder to points in the latent space then yields simulated data with possibly novel properties.

The objective function conventionally employed to optimize the encoder and decoder parameters, is obtained by considering, where $\mathbb{E}_{q_{\vec{\phi}}(\vec{z}|\vec{x})}$ indicates an average over the encoder output (typically for all samples in $X$),

$$\log p_{\vec{\theta}}(\vec{x}) = \mathbb{E}_{q_{\vec{\phi}}(\vec{z}|\vec{x})}\left[\log\left(\frac{p_{\vec{\theta}}(\vec{x},\vec{z})}{p_{\vec{\theta}}(\vec{z}|\vec{x})} \cdot \frac{q_{\vec{\phi}}(\vec{z}|\vec{x})}{q_{\vec{\phi}}(\vec{z}|\vec{x})}\right)\right]$$

$$= \mathbb{E}_{q_{\vec{\phi}}(\vec{z}|\vec{x})}\left[\log\left(\frac{p_{\vec{\theta}}(\vec{x},\vec{z})}{q_{\vec{\phi}}(\vec{z}|\vec{x})}\right)\right] + \mathbb{E}_{q_{\vec{\phi}}(\vec{z}|\vec{x})}\left[\log\left(\frac{q_{\vec{\phi}}(\vec{z}|\vec{x})}{p_{\vec{\theta}}(\vec{z}|\vec{x})}\right)\right] \quad (2)$$

The second, Kullback-Leibler (KL) divergence, term above is positive definite and effectively quantifies the difference between $q_{\vec{\phi}}(\vec{z}|\vec{x})$ and $p_{\vec{\theta}}(\vec{z}|\vec{x})$. The first term, often referred to as the evidence lower bound, equals according to the chain rule of probability,

$$\mathbb{E}_{q_{\vec{\phi}}(\vec{z}|\vec{x})}\left[\log\left(p_{\vec{\theta}}(\vec{x}|\vec{z})\right)\right] - \mathbb{KL}\left(q_{\vec{\phi}}(\vec{z}|\vec{x}) \parallel P(\vec{z})\right), \quad (3)$$

Eq.(3) bounds the logarithm of the likelihood of the data (the left hand side of Eq. (2)) as a result of the positivity of the KL divergence. Accordingly maximizing the evidence lower bound in Eq. (2) both enhances the marginal likelihood $p_{\vec{\theta}}(\vec{x})$ and decreases the distance between the encoder and the inverse of the decoder which further augments the statistical significance of the decoded samples and hence $p_{\vec{\theta}}(\vec{x})$. However, since the relative physical importance of the KL and evidence lower bound is problem-dependent, the KL term can be multiplied by a factor $\beta$ before optimization, yielding the so-called $\beta$-VAE procedures.[42–44]

Subsequently, $P(z)$ is generally set to $N(0, \mathbf{I})$, where $N$ and $\mathbf{I}$ represent respectively the normal distribution and a unit matrix with the latent space dimension. If the encoder is described by a second normal distribution of the form $q_{\vec{\phi}}(\vec{z}|\vec{x}) = \prod_i N(z_i; \mu_i, \sigma_i^2)$ where $\mu_i$ and $\log \sigma_i$ are determined by the neural network and are therefore $\vec{x}$-dependent the KL divergence becomes for a $k$-dimensional latent space

$$\mathbb{KL}\left( q_{\vec{\phi}}(\vec{z}|\vec{x}) \parallel P(\vec{z}) \right) = -\frac{1}{2} \sum_{i=1}^{k} \left(1 + \log \sigma_i^2(\vec{x}) - \sigma_i^2(\vec{x}) - \mu_i^2(\vec{x})\right) \tag{4}$$

To enable backpropagation the "reparameterization trick" duplicates the last layer of the encoder, and associates the $i$:th of the $k$ latent space components of one of the two branches with $\log \sigma_i^2$ in Eq.(4) and that of the second branch with $\mu_i$. The outputs from the two encoder branches are combined with the output of a normalized Gaussian random number generator such that $\vec{z} = \vec{\mu} + \vec{\sigma} \cdot \vec{\epsilon}$, with $\vec{\epsilon} = N(0, \mathbf{I})$, where the dot indicates an inner product, yielding a Gaussian probability density $p(\vec{\epsilon})$ for the latent variable coding. Since only $\vec{\epsilon}$ varies stochastically, $\mathbb{E}_{q_{\vec{\phi}}(\vec{z}|\vec{x})}$ can as desired be identified with $\mathbb{E}_{p(\vec{\epsilon})}$ such that as desired, gradients with respect to $\vec{\phi}$ commute with $\mathbb{E}_{p(\vec{\epsilon})}$. Further, after the reparameterization trick is implemented, the first term of Eq.(3) becomes simply

$$\mathbb{E}_{N(0,1)} \left[ \log \left( P\left( \vec{x} \middle| z_i = \mu_i(\vec{x}) + \epsilon_i \sqrt{\sigma_i(\vec{x})} \right) \right) \right] \tag{5}$$

which possesses finite and continuous gradients with respect to $\mu_i$ and $\sigma_i$.

Once the VAE is trained, a synthetic output realization is generated simply by assigning values to the latent variables $z_i$ and determining the resulting output distribution. If the inputs are differing discrete images, the resulting image in general interpolates between the characteristic elements of the sets. This procedure can be employed, for example, to construct artificial facial images from those of actual people.

*Modfied (Targeted Ouptut) Procedure:* This paper presents a modification of the above, standard VAE formalism that we term the "targeted output" procedure. In this method, the objective function of Eq.(2), is replaced by

$$\log p_{\vec{\theta}}(\vec{x}') = \mathbb{E}_{q_{\vec{\phi}}(\vec{z}|\vec{x})}\left[\log\left(\frac{p_{\vec{\theta}}(\vec{x}',\vec{z})}{p_{\vec{\theta}}(\vec{z}|\vec{x}')} \cdot \frac{q_{\vec{\phi}}(\vec{z}|\vec{x})}{q_{\vec{\phi}}(\vec{z}|\vec{x})}\right)\right] \tag{6}$$

where the vector $\vec{x}'$ contains the variable values of a fixed "target distribution" that incorporates and typifies the essential features of the class associated with the input data record $\vec{x}$. In the case considered here of the MNIST digits, the target distribution for each set of identical digits is somewhat arbitrarily chosen to be a single unrotated well-formed element chosen from the set; however, in general the reference distribution for a given class of images will typify the specific features of the class that are of greatest interest. Maximizing the term that replaces the evidence lower bound in Eq.(3) namely

$$\mathbb{E}_{q_{\vec{\phi}}(\vec{z}|\vec{x})}\left[\log\left(\frac{p_{\vec{\theta}}(\vec{x}',\vec{z})}{q_{\vec{\phi}}(\vec{z}|\vec{x})}\right)\right] \tag{7}$$

Then reduces the difference the Kullback-Leibler (KL) divergence

$$\mathbb{E}_{q_{\vec{\phi}}(\vec{z}|\vec{x})}\left[\log\left(\frac{q_{\vec{\phi}}(\vec{z}|\vec{x})}{p_{\vec{\theta}}(\vec{z}|\vec{x}')}\right)\right] \tag{8}$$

between $q_{\vec{\phi}}(\vec{z}|\vec{x})$ and $p_{\vec{\theta}}(\vec{z}|\vec{x}')$ distributed over all the rotated images corresponding to a given digit. That is, the encoder conditional probability distribution function that maps a rotated digit of a given class to a point in the latent vector space will approximate the conditional function for the decoder mapping from the unrotated target digit to the same latent vector space point. Additionally, the probability $p_{\vec{\theta}}(\vec{x}')$ is enhanced by the minimization. Especially since the output of the encoder is statistically perturbed by the reparameterization trick, the above requirements are optimally realized if the encoded data for all rotations of a given digit occupies a limited and isolated region in the low-dimensional latent vector space that is effectively an abstract representation that interpolates between the unrotated and rotated digits. That is, if the encoder were to map the rotated digit set of two different digits to the overlapping regions in latent vector space, since these points are randomly perturbed before the decoder layers are applied, the decoder could not correctly classify the digits. Rather, since the decoder maps the latent space points to an unrotated digit, the distance between the encoder and decoder is reduced if the encoder network learns a mapping to a region of latent vector space that effectively unrotates the digit up to the residual additional rotation that can be provided by the decoder. Consequently, the VAE not only learns the mapping between unrotated and rotated digits but also partitions latent space into distinct regions associated with the representations of all rotations of each digit where closely spaced points in latent vector space tend to represent nearly identically rotated digits. Of course, the fidelity of the results depends on the degree of training as well as the extent to which the distinguishing features associated with each set of rotated and/or translated images are present in the target distribution that is employed to characterize this set. The examination of the rotated MNIST digit set below however clearly illustrates the distinguishing features and potential applications of the technique.

**3. Comp utational Results:** As described in the appendix, the following analysis implements a modified version of the computer code and network architecture presented in Chapter 12 of [45]. The input data consists of the benchmark MNIST digit set with pixel values normalized to values between 0 and 1. This set contains a diverse collection of 70,000 handwritten digits discretized in a $28 \times 28$ point grid with 256 grayscale levels together with their associated labels. As in [45], the encoder consists of the input layer followed by two 2-dimensional convolution layers employing 32 and 64 filters with **strides** equal to 2, **same** padding and $3 \times 3$ filter functions. The resulting $7 \times 7$ filters are then flattened and inserted into a 16 element dense layer which is finally connected to the latent-space sampling layer. The matched decoder network then inverts this process by connecting the latent layer first to a $7 \cdot 7 \cdot 64$ element dense layer which is subsequently reshaped into a $7 \times 7 \times 64$ tensor and passed through the 2 dimensional transposed 64 and 32 filter layers and finally a 2-dimensional single $3 \times 3$ filter layer which condenses the information contained in the different filter outputs of the second transpose layer back into a single $28 \times 28$ matrix. The neural network is trained over 30 epochs at which point the results are effectively stable with respect to the epoch number. Except for the final layer, for which a **sigmoid** activation function outputs a number between 0 and 1 at each node only **relu** activations were employed. The calculation further incorporated an **Adam** optimizer with a batch size of 128.

To benchmark subsequent results for rotated digits, Figure 1 shows the distribution in the 2-dimensional latent vector space associated with the MNIST digits after training, where the digit labels are distinguished by grey level such that the latent space locations of the zero digits are displayed as black dots while the positions of the 9 digits are white dots (not visible). The left diagram corresponds to the result of the standard algorithm in which the loss function contains the cross entropy between the output distribution and the input distribution while for the right diagram the cross entropy employs instead fixed target output distributions that are well-formed digits selected from the MNIST dataset (the digits chosen to represent 0 through 9 are respectively digits 56, 102, 25, 50, 26, 175, 62, 15, 46 and 45 of the set). While the separation of the regions associated with different digits in latent vector space is clearly enhanced by the fixed target distributions, the inherent tradeoff inherent in this approach is evident in Figure 2. This figure displays a $30 \times 30$ array of the output patterns generated by applying the decoder to evenly spaced latent space points between -3 and +3 in each of the two coordinate directions where again the left and right figures are associated with VAE:s trained with the standard and fixed output distributions, respectively. As expected, for fixed targets, the patterns at the interface between regions of different digits are superpositions of rather than interpolations between the target distributions. Note that while Figure 1 and Figure 2 display similar content, the coordinate axes directions differ.

The significance of the fixed target procedure is apparent for input data such as that of Figure 3 in which the input MNIST digits are randomly rotated. Since a digit rotated through a large angle differs considerably from the unrotated digit, for the standard VAE technique the digit distributions in latent vector space are superpositions of the distributions associated with the digits at each rotation angle. Except for the rotationally symmetric zero digit, these typically overlap as evident from the left diagram of Figure 4. In contrast, the targeted output procedure yields the right figure of Figure 4 for which the latent space regions associated with each digit are well delineated. The associated decoder outputs for the evenly spaced set of latent grid points employed in Figure 2, are shown in Figure 5. The dominant output patterns are the circularly symmetric zero digit and a line resembling the digit 1 in various orientations, which presumably results from the average over all digits at each angular displacement.

The improvement in the latent space separation of the digits afforded by the targeted output procedure can also be visualized by applying the t-SNE algorithm. This method projects the latent space regions associated with the digits for any latent space dimension corresponding onto a compact two-dimensional surface such that the regions occupied by each digit remain contiguous and separated from those of the other digits. A comparison of the t-SNE patterns in the case of rotated digits yields the diagrams of Figure 6, where again the left and right diagrams again refer to the standard and targeted output procedures, respectively. As expected, the t-SNE method locates separated regions corresponding to different digits for the targeted output technique but is only able to discriminate the zero-digit region in the case of the standard method.

Since random rotations are associated with an additional degree of freedom, groups of rotated digits are typically better differentiated in latent vector spaces of higher dimensions. Figure 7 accordingly displays the distribution of the randomly rotated input digits in a three-dimensional latent vector space, with the left and right diagrams respectively corresponding to the standard and targeted output VAE procedures. While the additional latent space dimension increases the separation among the sets of rotated digits in both methods, the enhancement afforded by the targeted output technique is again far greater. The three and two-dimensional results can be more directly compared by employing the t-SNE procedure to map the three-dimensional results of Figure 7 onto the two-dimensional manifolds displayed in Figure 8. As expected, the digits occupy more isolated latent space volumes in both the standard and targeted output procedures in a three- compared to a two-dimensional latent space. However, the improvement afforded by the targeted output algorithm is far greater than that resulting from increasing the dimensionality in the standard procedure.

The diagrams of Figure 9 employ a 10-dimensional latent vector space for which each digit could in theory be associated with a different orthogonal latent vector direction. The left and middle of these respectively apply the VAE to randomly rotated digits with the standard and fixed output procedures while the right diagram employs the standard procedure with non-rotated input digits. Interestingly, the degree of isolation among the different digit clusters in the latent vector space for the targeted output distribution (the middle diagram) is here nearly identical to that associated with unrotated digits for the standard method (the right diagram).

Subsequently, Figure 10 displays the two-dimensional latent vector space distribution for randomly rotated digits obtained with the standard method and $\beta$ factors of 0.1 and 10, respectively. The effect of the $\beta$ factor is here seen to be rather minimal as would be expected from physical origin of the overlapping distributions. While not shown, the impact of changing the $\beta$ factor is somewhat more noticeable in the case of the targeted output method; however, the separation between different digit regions in this case is quite satisfactory for all values of $\beta$.

The final figure, Figure 11, displays, for a three-dimensional latent vector space, the output patterns generated by the targeted output VAE procedure for latent data space points in the immediate vicinity of the latent data space location of the reference pattern for the number 8. To obtain this result, the latent space location was first identified by passing the reference pattern through the encoder. Subsequently, a cube with side length 0.2 centered on the latent space location was constructed. To generate a distribution of sample points, the MNIST digits were then again randomly rotated and passed through the encoder. It was found that for the input instances that mapped to points within the cube 1, 2, 326 and 1 of these corresponded to the digits 1, 3, 8 and 9, respectively. This agrees with the qualitative similarity of the digits 3 and 9 with the number 8. A random sampling of the digits that were mapped to latent space points inside the cube are displayed in Figure 11 and clearly demonstrate that the modified VAE is able to discriminate the elements of the digit class independent of rotation angle. A complete procedure for discriminating among the digits would require the determination of decision boundaries between the latent space regions corresponding to the different digits and is accordingly beyond the scope of this article. However, the ability of the procedure to distinguish in a simple fashion transformed members of a given class by identifying patterns that are mapped by the encoder to the vicinity of the latent space point of the pattern employed to characterize the class elements should by itself be of considerable importance in certain contexts.

**4. Conclusions and Future Directions:** Randomly rotated digits are in general mapped by a VAE to overlapping regions of the latent variable space because of the large dissimilarities among the patterns associated with a given digit for differing rotation angles (with the obvious exception of the zero digit). While the degree of coincidence is reduced for high-dimensional latent vector spaces, it was here instead largely eliminated by specifying a certain non-rotated characteristic target distribution for each rotated digit at the VAE output and redefining the objective function such that it quantifies the deviation of the VAE output from this target distribution. This targeted output technique then displays a significant discrimination in latent space among digits even for two-dimensional latent spaces. Such behavior sets the procedure as well apart from other generative algorithms such as restricted Boltzmann machines [46] and principal component analysis, which generally require large latent space dimensions to achieve high accuracy. [47]

The fixed target procedure could potentially be adapted to a variety of applications. In pattern recognition each input data record that is associated with a certain set of salient features could be mapped onto a fixed target distribution that embodies these characteristics. After the VAE is trained, the VAE could classify new data records by mapping the records to their corresponding regions of the latent vector space.

On a more fundamental level, the results of this paper suggest the existence of a tradeoff between the ability of a VAE to discriminate between different input classes and the degree to which it can smoothy interpolate between the elements of these classes. That is, by training the VAE with randomly rotated input data on fixed patterns rather than on the actual input data, the regions between those associated with different digits in the latent vector space map after decoding to superpositions rather than interpolations of the target distributions. However, since the latent space representation of the standard VAE model is in any case unable to resolve different classes of data for complex problems such as the fully rotated digit example considered here unless the algorithm is appropriately modified, this tradeoff is often largely irrelevant.

**Acknowledgements:** The Natural Sciences and Engineering Research Council of Canada (NSERC) is acknowledged for financial support. [grant number RGPIN-03907-2020]

**Biography:** David Yevick (Ph.D. 1979, F. OSA, IEEE, APS) is a professor of physics at the University of Waterloo having been previously at Queen's University (Kingston), Penn State University ,Lund University and the Institute of Optical Research, Stockholm. He has published over 200 articles in optical communications, physics and computational methods.

**Appendix:** The program employed to generate the results of this paper effectively follows the VAE code on Chapter 12 of [45] with the following critical modifications. First, the MNIST samples employed as target digits are stored in the tensor **myData** as follows:

```
myData = np.array( [ (mnist_digits[56]),
         (mnist_digits[102]),
         (mnist_digits[25]),
         (mnist_digits[50]),
         (mnist_digits[26]),
         (mnist_digits[175]),
         (mnist_digits[62]),
         (mnist_digits[15]),
         (mnist_digits[46]),
         (mnist_digits[45])                                      ]                           );
```

The essential change in the VAE code is then the replacement of the standard loss calculation for each of the data samples, where **reconstruction** is the decoder output for the sample and **x_train** is the MINST sample digit pattern, namely,

```
  reconstruction_loss = tf.reduce_mean(
    tf.reduce_sum(
      keras.losses.binary_crossentropy(x_train, reconstruction),
      axis = (1, 2)
    )
  )
```

by the code, where **y_train** is the integer label associated with **x_train.**

```
reconstruction_loss = tf.reduce_mean(
  tf.reduce_sum(
    keras.losses.binary_crossentropy(myData[y_train], reconstruction),
    axis = (1, 2)
  )
)
```

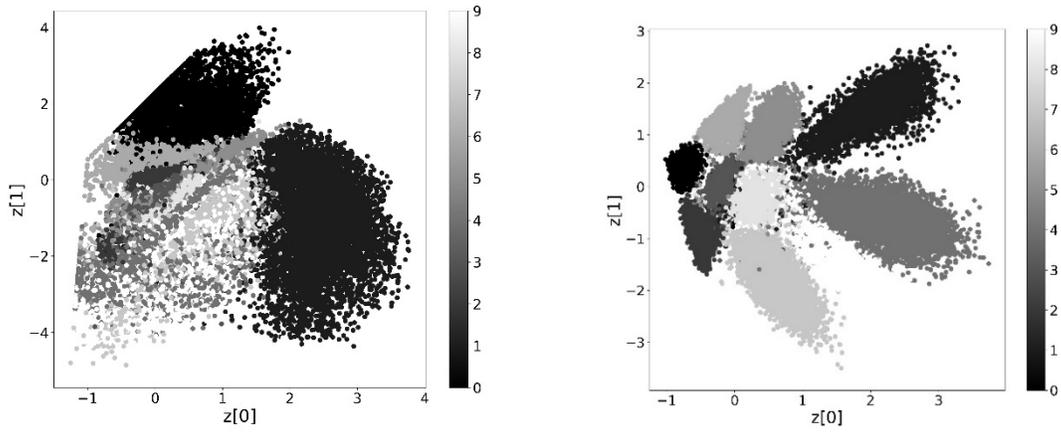

*Figure 1*: *The 2 dimensional latent variable space distribution generated by the unrotated MNIST digit set with the standard (left plot) and fixed target (right plot) VAE procedures. The digit values are shaded as indicated in the colorbar.*

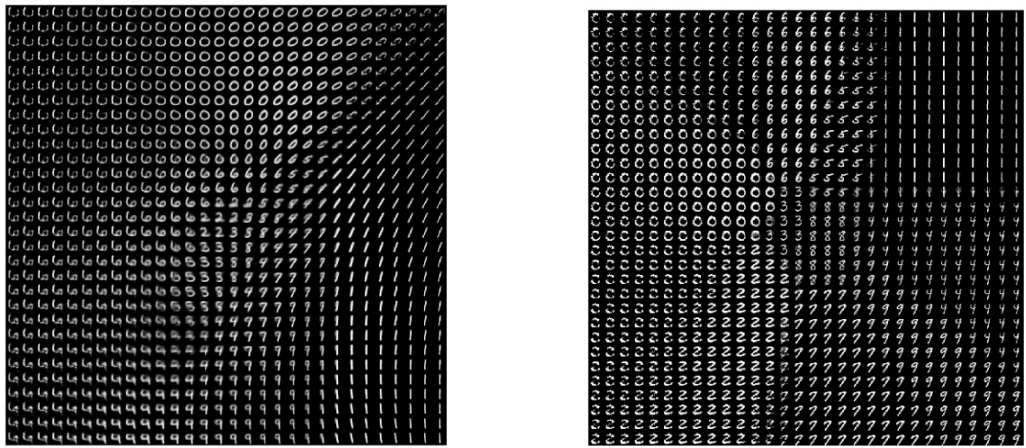

*Figure 2*: The neural network output corresponding to different 2 -dimensional latent space points for the unrotated MNIST digit set with the standard (left plot) and fixed target (right plot) VAE procedures

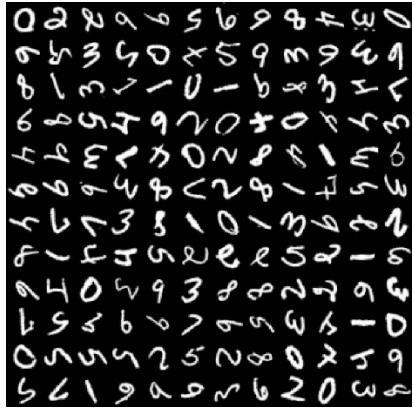

*Figure 3*: A sample of the rotated MNIST digits

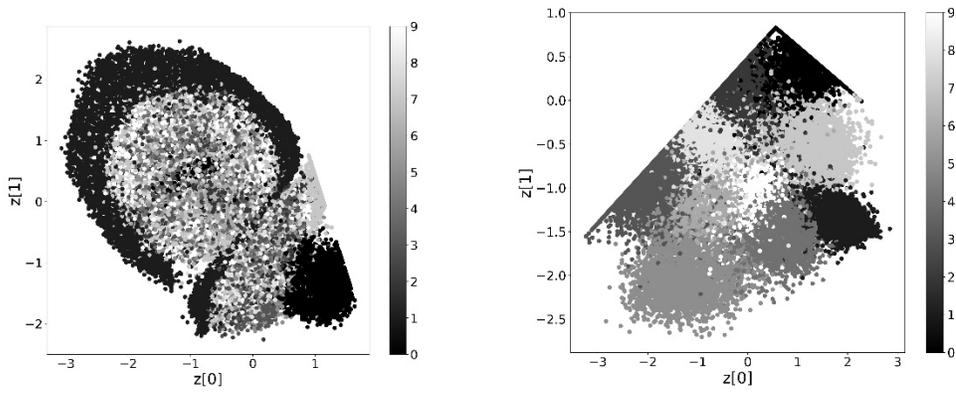

*Figure 4*: The 2 dimensional VAE latent variable space for the randomly rotated MNIST digit set with the standard procedure (left plot) and for fixed target distributions (right plot)

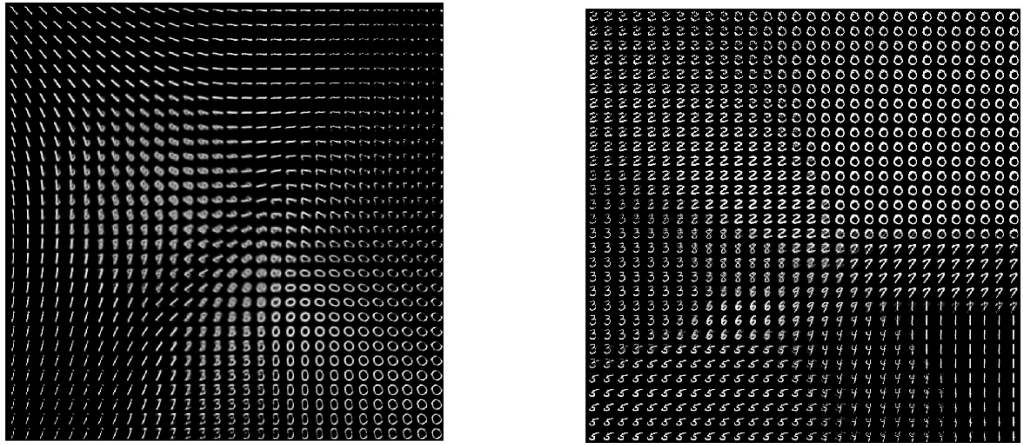

*Figure 5*: The output patterns generated by a rectangular grid of points in a 2-dimensional latent space for *the randomly rotated MNIST digit set with the standard procedure (left plot) and for fixed target distributions (right plot)*

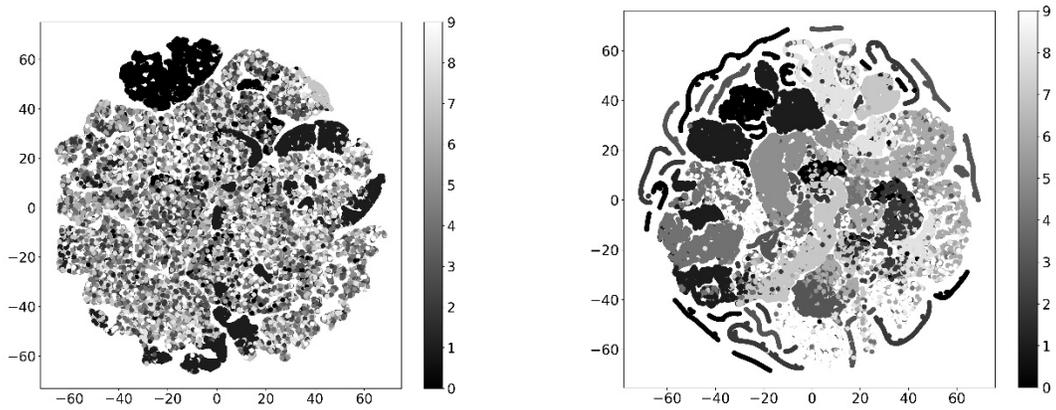

*Figure 6*: As in the previous figure, but for the corresponding t-SNE results

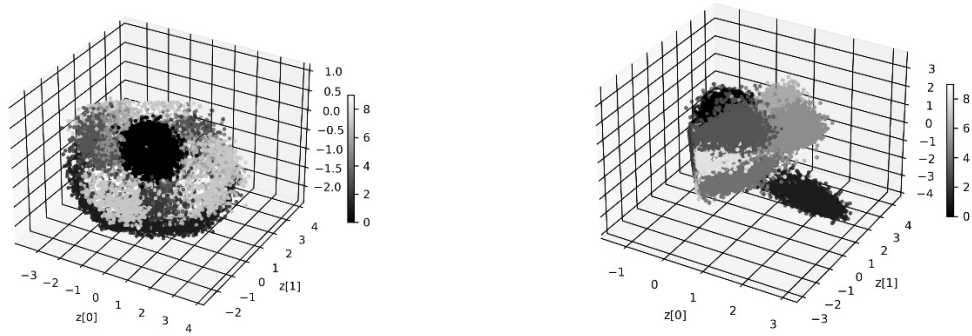

*Figure 7*: The 3-dimensional latent space *for the randomly rotated MNIST digit set with the standard procedure (left plot) and for fixed target distributions (right plot)*

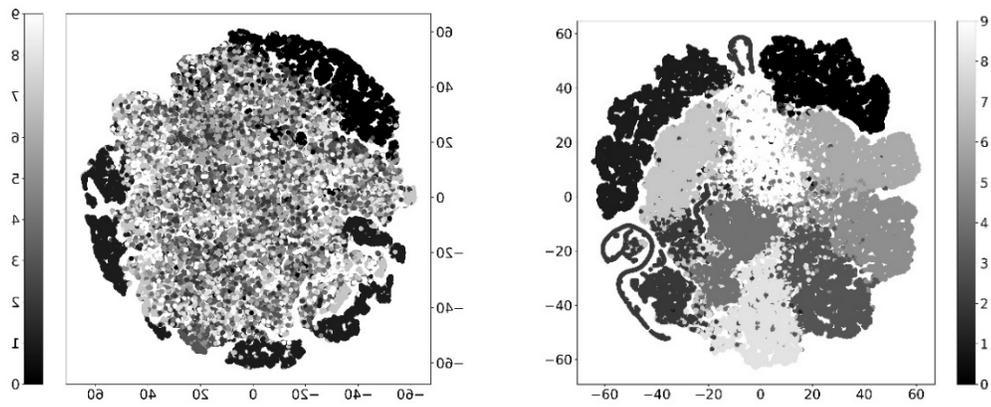

*Figure 8*: The t-SNE patterns associated with the previous figure.

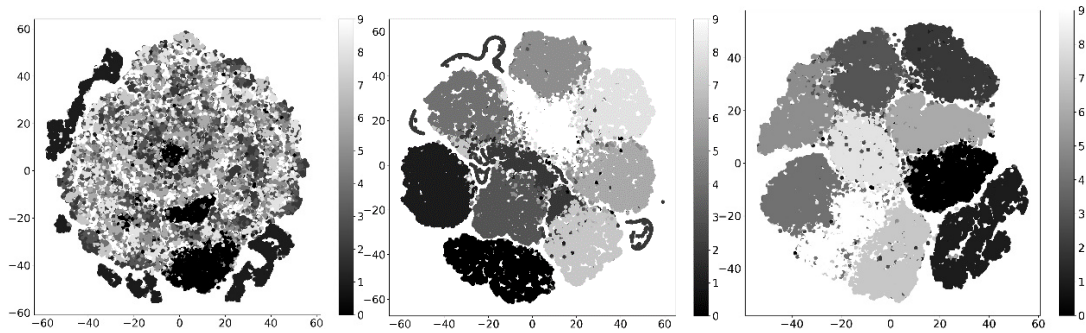

*Figure 9*: The t-SNE patterns for a ten dimensional latent space generated by randomly rotated digits with the standard procedure (left plot) and fixed target distributions (middlet plot) and for unrotated digits with the standard procedure (right plot).

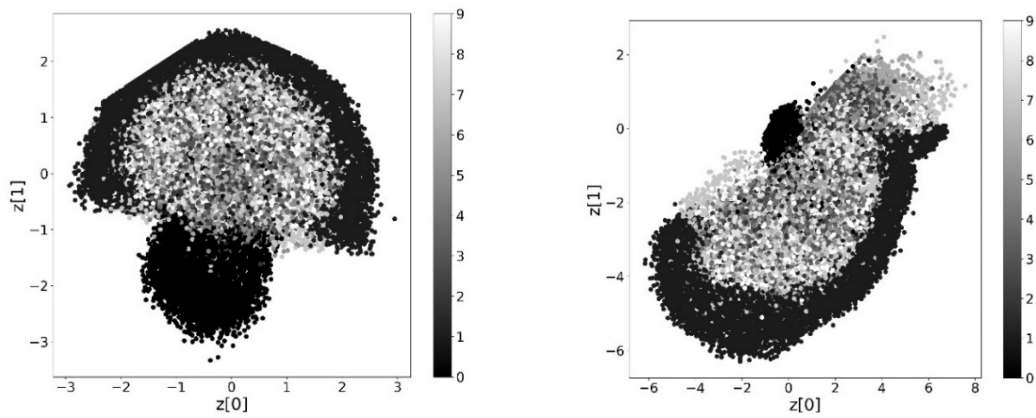

*Figure 10*: The 2-dimensional latent space distribution and t-SNE pattern for the randomly rotated MINST digit set with KL amplification factors of 10 and 0 (left and right pictures respectively).

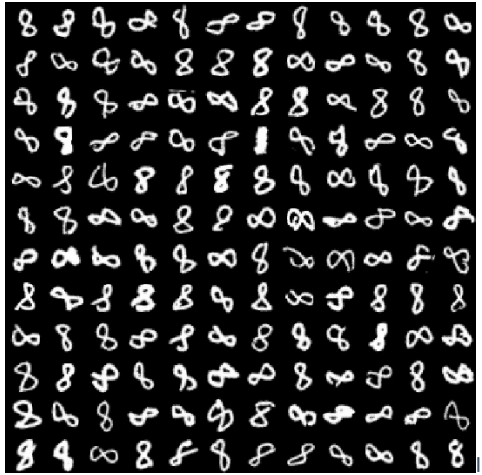

*Figure 11*: The output patterns generated by latent space points in the vicinity of the latent space location of the reference pattern for the number 8 and a three-dimensional latent vector space.